\def\year{2019}\relax
\documentclass[letterpaper]{article} 
\usepackage{aaai19}  
\usepackage{times}  
\usepackage{helvet}  
\usepackage{courier}  
\usepackage{url}  
\usepackage{graphicx}  
\usepackage{algorithm}  
\usepackage{algpseudocode}  
\usepackage[colorlinks,
            linkcolor=black,
            anchorcolor=black,
            citecolor=black,
            urlcolor=black
            ]{hyperref}
\usepackage{amsmath}  
\usepackage{color}  

\begin{document}
%
\title{A Unified Framework of DNN Weight Pruning and Weight Clustering/Quantization Using ADMM}

\author{\small{Shaokai Ye$^{1*}$, Tianyun Zhang$^{1*}$, Kaiqi Zhang$^{1}$, Jiayu Li$^1$, jiaming xie$^2$, Yun Liang$^2$, Sijia Liu$^3$, Xue Lin$^4$ \& Yanzhi Wang$^4$}\\ 
1. Syracuse University, USA \\
\texttt{\{sye106,tzhan120,jli221,kzhang17\}@syr.edu} \\
2. Peking University, China\\
3. MIT-IBM Watson AI Lab, IBM Research\\
4. Northeastern University, USA  \\ 
$^*$Equal Contribution\\
}

\maketitle
\begin{abstract}
Many model compression techniques of Deep Neural Networks (DNNs) have been investigated, including weight pruning, weight clustering and quantization, etc. Weight pruning leverages the redundancy in the number of weights in DNNs, while weight clustering/quantization leverages the redundancy in the number of bit representations of weights. They can be effectively combined in order to exploit the maximum degree of redundancy. However, there lacks a systematic investigation in literature towards this direction.

In this paper, we fill this void and develop a unified, systematic framework of DNN weight pruning and clustering/quantization using Alternating Direction Method of Multipliers (ADMM), a powerful technique in optimization theory to deal with non-convex optimization problems. Both DNN weight pruning and clustering/quantization, as well as their combinations, can be solved in a unified manner. For further performance improvement in this framework, we adopt multiple techniques including iterative weight quantization and retraining, joint weight clustering training and centroid updating, weight clustering retraining, etc. The proposed framework achieves significant improvements both in individual weight pruning and clustering/quantization problems, as well as their combinations. For weight pruning alone, we achieve 167$\times$ weight reduction in LeNet-5, 24.7$\times$ in AlexNet, and 23.4$\times$ in VGGNet, without any accuracy loss. For the combination of DNN weight pruning and clustering/quantization, we achieve 1,910$\times$ and 210$\times$ storage reduction of weight data on LeNet-5 and AlexNet, respectively, without accuracy loss. Our codes and models are released at the link \url{http://bit.ly/2D3F0np}.
\end{abstract}

\section{Introduction}
Despite the significant success and wide applications, Deep Neural Networks (DNNs) have increasing model sizes and associated computation and storage overheads. 
DNN model compression techniques have been widely investigated, including \emph{weight pruning} \cite{han2015,wen2016learning,dai2017,guo2016dynamic}, \emph{weight clustering and quantization} \cite{DeepCompression,park2017weighted,zhou2017incremental,Leng2017}, low-rank approximation \cite{cheng2015exploration,sindhwani2015structured,zhao2017theoretical}, etc.

A pioneering work on weight pruning is the iterative pruning method \cite{han2015}, which successfully prunes 12$\times$ weights in LeNet-5 (MNIST dataset) and 9$\times$ weights in AlexNet (ImageNet dataset), without accuracy degradation. 
The limitations of this work include (i) the limited capability in weight pruning in convolutional (CONV) layers, the most computationally intensive component in DNNs, and (ii) the irregularity in weight storage after pruning. To overcome these limitations, multiple recent work have extended to (i) use more sophisticated algorithms for higher pruning ratio \cite{zhang2018systematic,ye2018progressive,dai2017,guo2016dynamic}, (ii) strike a balance between higher pruning ratio and lower accuracy degradation \cite{yang2016}, and (iii) incorporate regularity in weight pruning and storage to facilitate hardware implementations \cite{wen2016learning,wen2017coordinating,zhang2018adam}.

Weight clustering and quantization are equally important, if not more, in DNN model compression. Weight clustering is different from quantization: the former requires the weights to be grouped into a predefined number of clusters, and weights within a cluster are the same; the latter requires the weights to take pre-defined, fixed values. In fact, weight quantization is a special type of the general weight clustering. Due to its flexibility, weight clustering will result in higher accuracy and/or compression ratio than quantization. On the other hand, weight quantization, especially equal-distance quantization, is more hardware friendly compared to weight clustering (and weight pruning as well).
Both weight clustering and quantization will be considered in this paper in a unified way. 

Many research work are dedicated to weight clustering and quantization \cite{Leng2017,zhou2016dorefa}. The current work are mainly iterative, including back-propagation training assuming continuous weights and mapping procedure to discrete values. It is important to note that multiplications can even be eliminated through effective weight quantization \cite{Leng2017}, through quantization into binary weights, ternary weights ($-1$, 0, $+1$), or weight quantization into powers of 2 such as the DoReFa net \cite{zhou2016dorefa}. 

Weight pruning makes use of the redundancy in the number of weights in DNNs, whereas weight clustering/quantization exploits the redundancy in weight representations. These two sources of redundancy are largely independent with each other, which makes it desirable to combine weight pruning and clustering/quantization to make full exploitation of the degree of redundancy.
Despite some early heuristic investigation \cite{han2015,DeepCompression}, there lacks a systematic investigation on the best possible combination of DNN model compression techniques. This paper aims to overcome this limitation and shed some light on the highest possible DNN model compression through effective combinations.

This paper develops a unified, systematic framework of DNN weight pruning and weight clustering/quantization using \emph{Alternating Direction Method of Multipliers} (ADMM), a powerful technique in optimization to deal with non-convex optimization problems with potentially combinatorial constraints \cite{boyd2011,takapoui2017simple}. This framework is based on a key observation: both DNN weight pruning and weight clustering/quantization, as well as their combinations, can be solved in a unified manner using ADMM. In the solution, the original problem is decomposed into three (or two) subproblems, which are iteratively solved until convergence. Overall, the ADMM-based solution can be understood as a smart, dynamic regularization process in which the regularization target is dynamically updated in each iteration. As a result it can outperform the prior work on regularization \cite{wen2016learning} or projected gradient descent \cite{zhang2018learning}.

For further performance improvement in ADMM-based weight clustering/quantization, we propose multiple techniques including iterative weight quantization and retraining procedure, joint weight clustering training and adaptive centroid updating, weight clustering retraining process, etc.
The proposed unified framework using ADMM outperforms prior work in two aspects. First, for individual weight pruning and clustering/quantization methods, the proposed ADMM method outperforms prior work. For instance, we achieve 167$\times$ weight reduction in LeNet-5, 24.7$\times$ in AlexNet, and 23.4$\times$ in VGGNet, without any accuracy loss, which clearly outperform prior arts. Second, for the joint DNN weight pruning and clustering/quantization, we achieve 1,910$\times$ and 210$\times$ storage reduction of weight data on LeNet-5 and AlexNet, respectively, without accuracy loss. These results significantly outperform the state-of-the-art results. Our codes and models are released at the link \url{http://bit.ly/2D3F0np}.

\section{Discussions on the Combination of DNN Weight Pruning and Clustering/Quantization}
As discussed before, weight pruning can be largely combined with weight clustering/quantization, thereby making full exploitation of the degree of redundancy. Preliminary work \cite{DeepCompression} in this direction uses a combination of iterative weight pruning and K-means clustering methods. It simultaneously achieves 9$\times$ weight pruning in AlexNet, and uses 8-bit CONV layer clustering and 5-bit FC layer clustering. This work does not target hardware implementation and only focuses on weight clustering.

When comparing with weight clustering/quantization, weight pruning can often result in a higher compression ratio of DNN \cite{DeepCompression}. This is because of two reasons. First, there is often higher degree of redundancy in the number of weights than the number of bits for weight representations. 
For weight clustering/quantization, for a single bit reduction in weight representation, the imprecision can be perceived to be doubled. This difficulty is not faced by weight pruning methods.
Second, moderate weight pruning can often result in an increase in accuracy (by up to 2\% in AlexNet in our ADMM framework), thereby resulting in a higher margin for further weight reduction. This effect, however, is not observed in weight clustering/quantization. 
As a result, weight pruning is often prioritized over weight clustering/quantization despite the effect of irregular weight storage and associated hardware implementation overhead in the former method. 

In the prior work, there lacks a systematic investigation on the best possible combination of DNN weight pruning and weight clustering/quantization methods. In this paper we fill this void in order to make the full exploitation of the degree of redundancy. We provide formulation that can both perform ADMM-based weight pruning and clustering/quantization simultaneously, or give priority to weight pruning.

Fig. 1 shows an illustrative process about weight quantization after weight pruning. Given that a $4\times 4$ weight matrix after pruning will be quantized on Fig. 1 (a), we use 2-bit for quantization and the interval is 0.5. 
Then the quantization levels become $\{-1,-0.5,0.5,1\}$ without 0 because 0 represents pruned weights. Fig. 1 (b) shows the quantized weights, and Fig. 1 (c) displays the values that are actually stored in hardware along with the interval value 0.5.

\begin{figure}
\begin{center}
\includegraphics[width=\linewidth]{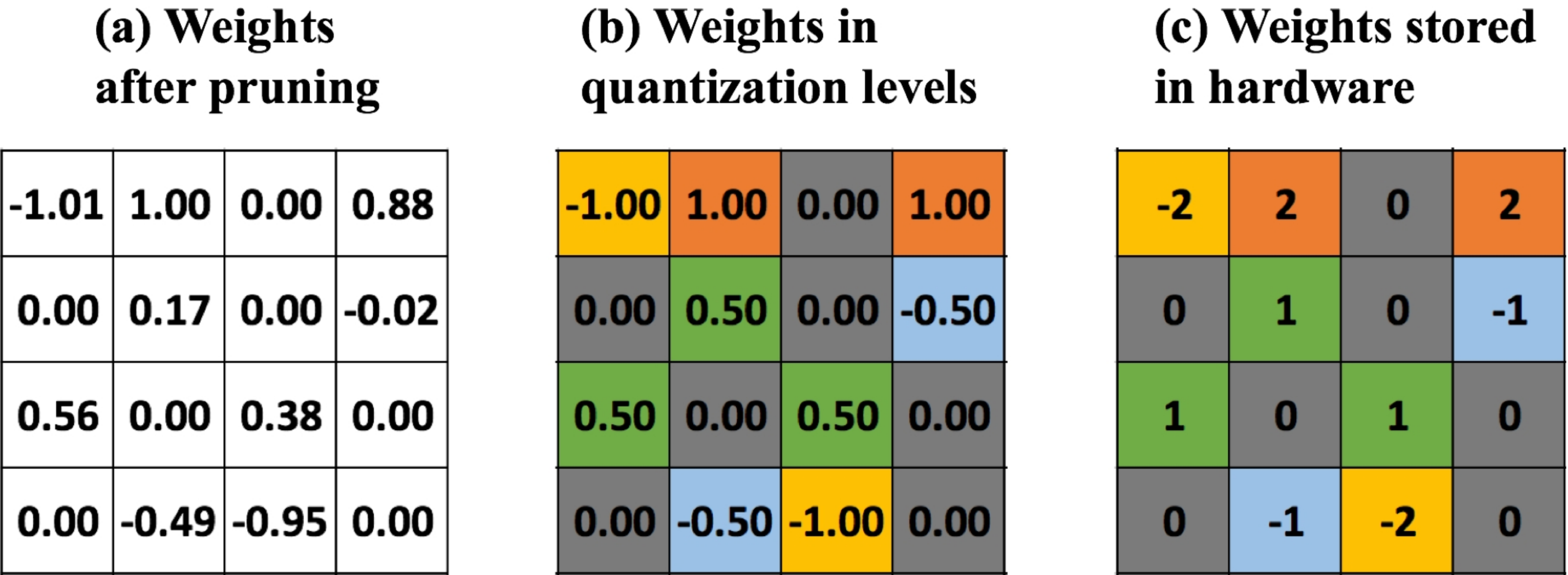}
\caption{Description of weight quantization after weight pruning (the interval $q_i$ equal to 0.5).}
\label{fig:example}
\end{center}
\end{figure}

Fig. 2 shows the illustrative weight clustering process after weight pruning, which is different from weight quantization. Given the same $4\times 4$ weight matrix after pruning, we also use 2-bit for weight clustering. Again 0 is not considered because the associated weights are already pruned. Fig. 2 (b) shows the weights after clustering, along with the centroid values for the 4 clusters shown in Fig. 2 (c). Different from weight quantization, the centroid values are flexible in the weight clustering process.

\begin{figure}
\begin{center}
\includegraphics[width=\linewidth]{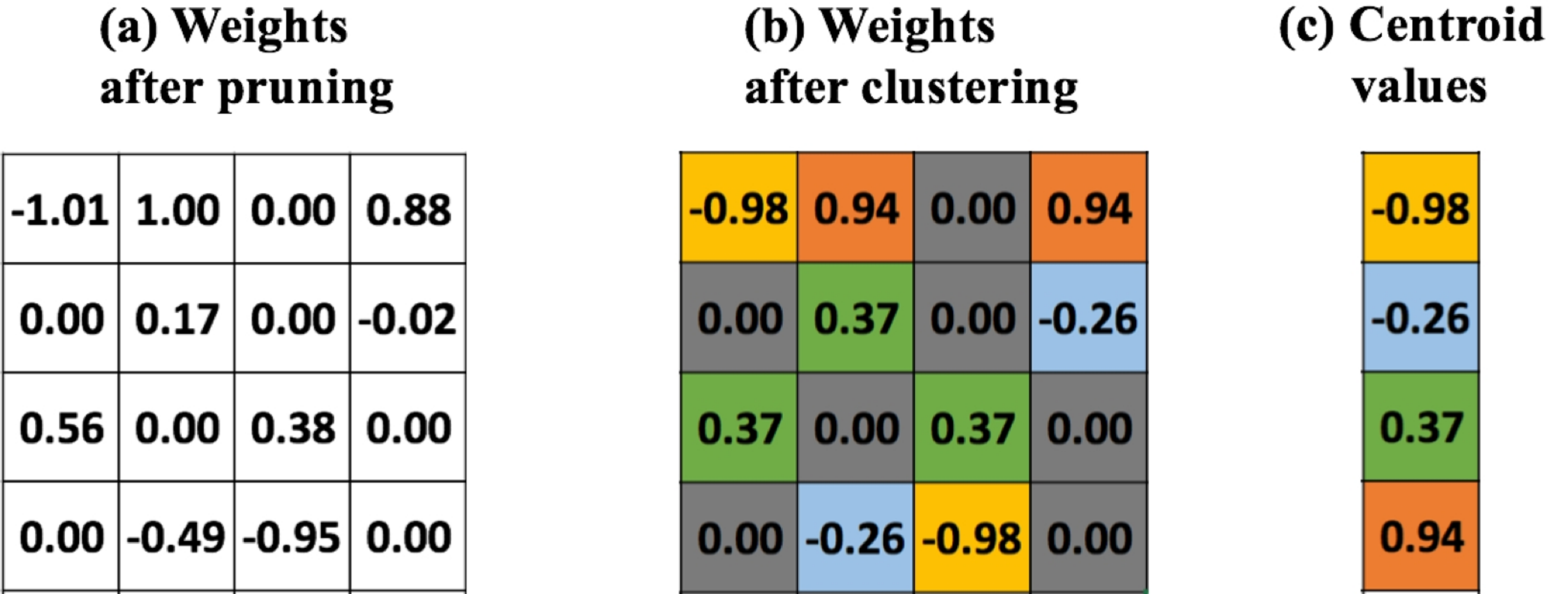}
\caption{Description of weight clustering after weight pruning (along with centroid values).}
\label{fig:example_cluster}
\end{center}
\end{figure}

\section{The Unified Framework of ADMM based Weight Pruning and Clustering/Quantization}

\subsection{Background of ADMM}
Consider a non-convex optimization problem that is difficult to solve directly. The ADMM method decomposes it into two subproblems that can be solved separately and efficiently. For example, the optimization problem
\begin{equation}
   \min_{\bf{x}}\ \ f({\bf{x}})+g({\bf{x}})
\end{equation}
lends itself to the application of ADMM if $f(\bf{x})$ is differentiable and $g(\bf{x})$ has some structure such as $L_0$/$L_1$-norm or the indicator function of a constraint set. The problem is first re-written as
\begin{equation}
\begin{aligned}
& \min_{\bf{x},\bf{z}}
& & f(\textbf{x})+g(\textbf{z}),
\\ & \text{subject to}
& & \bf{x}=\bf{z}.
\end{aligned}
\end{equation}
By using augmented Lagrangian \cite{boyd2011}, this problem is decomposed into two subproblems on $\bf{x}$ and $\bf{z}$. The first is $\min_{\bf{x}} f({\bf{x}})+q_1(\bf{x})$, where $q_1(\bf{x})$ is a quadratic function. As a result, the complexity of solving subproblem 1 (e.g., via stochastic gradient descent) is the same as minimizing $f(\bf{x})$. Subproblem 2 is $\min_{\bf{z}} g({\bf{z}})+q_2(\bf{z})$, where $q_2(\bf{z})$ is quadratic. When $g$ has special structure, exploiting the properties of $g$ allows this problem to be solved analytically and optimally. In this way, we can solve the problem via ADMM that is difficult to solve directly.

\subsection{Problem Formulation}
Consider an $N$-layer DNN, the collections of weights and biases of the $i$-th layer are respectively denoted by ${\bf{W}}_{i}$ and ${\bf{b}}_{i}$; The loss function of the $N$-layer DNN is denoted by $f \big( \{{\bf{W}}_{i}\}_{i=1}^N, \{{\bf{b}}_{i} \}_{i=1}^N \big)$.

When we combine DNN weight pruning with clustering or quantization, the overall problem is defined by
\begin{equation}
\begin{aligned}
& \underset{ \{{\bf{W}}_{i}\},\{{\bf{b}}_{i} \}}{\text{minimize}}
& & f \big( \{{\bf{W}}_{i}\}_{i=1}^N, \{{\bf{b}}_{i} \}_{i=1}^N \big),
\\ & \text{subject to}
& & {\bf{W}}_{i}\in {\bf{S}}_{i}, \; {\bf{W}}_{i}\in {\bf{S}}_{i}^{'}, \; i = 1, \ldots, N.
\end{aligned}
\end{equation}
The set ${\bf{S}}_{i}$ reflects the constraint for the weight pruning problem, i.e., ${\bf{S}}_{i} =
\{ \text{the number of nonzero elements is less}$ $\text{than or equal to} ~\alpha_i \}$, where $\alpha_i$ is the desired number of weights after pruning in the $i$-th layer. When we combine weight pruning with weight clustering, ${\bf{S}}_{i}^{'}
=
\{{\bf{W}}_i\mid \text{the weights in }{\bf{W}}_i ~ \text{contain no more than $M_i$ different values}\}$. When we combine weight pruning with weight quantization, ${\bf{S}}_{i}^{'}= \{{\bf{W}}_i\mid \text{the weights in } {\bf{W}}_i \text{ only take values from the }$ $\text{set }\{Q_1, Q_2, \cdots, Q_{M_i} \}\}$. 
Here the $Q$ values are \emph{quantization levels}, and we consider \emph{equal-distance quantization} (the same distance between quantization levels) to facilitate hardware implementations.
Besides, $M_i=2^n$, and $n$ is the number of bits we use for weight clustering or quantization. 

Both constraints ${\bf{S}}_{i}$ and ${\bf{S}}_{i}^{'}$ need to be satisfied simultaneously in the joint problem of DNN weight pruning and weight clustering/quantization. In this way we can make sure that most of the DNN weights are pruned (set to zero), while the remaining weights are clustered/quantized.

\subsection{The Unified ADMM-based Framework}
To apply ADMM, we define indicator functions to incorporate the combinatorial constraints into objective function. The indicator functions are \begin{eqnarray*}g_{i}({\bf{W}}_{i})=
\begin{cases}
 0 & \text { if } {\bf{W}}_{i}\in {\bf{S}}_{i}, \\ 
 +\infty & \text { otherwise, }
\end{cases}
\end{eqnarray*}
\begin{eqnarray*}h_{i}({\bf{W}}_{i})=
\begin{cases}
 0 & \text { if } {\bf{W}}_{i}\in {\bf{S}}_{i}^{'}, \\ 
 +\infty & \text { otherwise, }
\end{cases}
\end{eqnarray*}
for $i = 1, \ldots, N$.

We then incorporate auxiliary variables ${\bf{Z}}_{i}$ and ${\bf{Y}}_{i}$, and rewrite the original problem (3) as 
\begin{equation}
\label{admm_form}
\begin{aligned}
& \underset{ \{{\bf{W}}_{i}\},\{{\bf{b}}_{i} \}}{\text{minimize}}
& & f \big( \{{\bf{W}}_{i} \}_{i=1}^N, \{{\bf{b}}_{i} \}_{i=1}^N \big)+\sum_{i=1}^{N} g_{i}({\bf{Z}}_{i})+\sum_{i=1}^{N} h_{i}({\bf{Y}}_{i}),
\\ & \text{subject to}
& & {\bf{W}}_{i}={\bf{Z}}_{i}, \; {\bf{W}}_{i}={\bf{Y}}_{i}, \; i = 1, \ldots, N.
\end{aligned}
\end{equation}

Through ADMM \cite{boyd2011}, problem (\ref{admm_form}) can be decomposed into three subproblems. The overall problem of weight pruning and clustering/quantization is solved through solving the subproblems iteratively until convergence.
The first subproblem is
\begin{equation}
\begin{aligned}
\label{4}
 \underset{ \{{\bf{W}}_{i}\},\{{\bf{b}}_{i} \}}{\text{minimize}}
\ \ \ & f \big( \{{\bf{W}}_{i} \}_{i=1}^N, \{{\bf{b}}_{i} \}_{i=1}^N \big)+\sum_{i=1}^{N} \frac{\rho_{i}}{2}  \| {\bf{W}}_{i}-{\bf{Z}}_{i}^{k}+{\bf{U}}_{i}^{k} \|_{F}^{2} \\& +  \sum_{i=1}^{N} \frac{\rho_{i}}{2}  \| {\bf{W}}_{i}-{\bf{Y}}_{i}^{k}+{\bf{V}}_{i}^{k} \|_{F}^{2}, \\
\end{aligned}
\end{equation}
where ${\bf{U}}_{i}^{k}$ and ${\bf{V}}_{i}^{k}$ are the dual variables updated in each ADMM iteration. The first term in (\ref{4}) is the differentiable loss function of the DNN, while the other terms are quadratic terms and they are differentiable and convex. As a result, this subproblem can be solved by stochastic gradient descent (e.g., the ADAM algorithm \cite{kingma2014adam}) and the complexity of solving this subproblem is the same as training of the original DNN.

The second subproblem is
\begin{equation}
\label{5}
 \underset{ \{{\bf{Z}}_{i} \}}{\text{minimize}}
\ \ \ \sum_{i=1}^{N} g_{i}({\bf{Z}}_{i})+\sum_{i=1}^{N} \frac{\rho_{i}}{2} \| {\bf{W}}_{i}^{k+1}-{\bf{Z}}_{i}+{\bf{U}}_{i}^{k} \|_{F}^{2}. \\
\end{equation}
As we mentioned before, $g_{i}(\cdot)$ is the indicator function of ${\bf{S}}_{i}$, thus the analytical solution of problem (\ref{5}) is 
\begin{equation}
\label{6}
  {\bf{Z}}_{i}^{k+1} = {{\bf{\Pi}}_{{\bf{S}}_{i}}}({\bf{W}}_{i}^{k+1}+{\bf{U}}_{i}^{k}),
\end{equation}
where ${{\bf{\Pi}}_{{\bf{S}}_{i}}(\cdot)}$ is Euclidean projection of ${\bf{W}}_{i}^{k+1}+{\bf{U}}_{i}^{k}$ onto the set ${\bf{S}}_{i}$. In DNN weight pruning, $\alpha_i$ is the desired number of weights after pruning in the $i$-th layer. The Euclidean projection is to keep $\alpha_i$ elements in ${\bf{W}}_{i}^{k+1}+{\bf{U}}_{i}^{k}$ with the largest magnitude and set the rest to be zero \cite{boyd2011,zhang2018systematic}.

The third subproblem is 
\begin{equation}
\label{new_ variable}
 \underset{ \{{\bf{Z}}_{i} \}}{\text{minimize}}
\ \ \ \sum_{i=1}^{N} h_{i}({\bf{Y}}_{i})+\sum_{i=1}^{N} \frac{\rho_{i}}{2} \| {\bf{W}}_{i}^{k+1}-{\bf{Y}}_{i}+{\bf{V}}_{i}^{k} \|_{F}^{2}. \\
\end{equation}
Similar to the second subproblem, the solution of problem (\ref{new_ variable}) is 
\begin{equation}
\label{new_projection}
  {\bf{Y}}_{i}^{k+1} = {{\bf{\Pi}}_{{\bf{S}}_{i}^{'}}}({\bf{W}}_{i}^{k+1}+{\bf{V}}_{i}^{k}),
\end{equation}

For the weight quantization problem, the quantization levels $Q_1$, $Q_2$, ..., $Q_{M_i}$ are fixed. In fact, weight quantization is a special type of clustering in which the clustering centroids are pre-determined and fixed. In weight quantization, the Euclidean projection is to map every element of ${\bf{W}}_{i}^{k+1}+{\bf{V}}_{i}^{k}$ to the quantization level (centroid) closest to that element.

For weight clustering, the centroids of the clusters can be updated dynamically, and the constraint is on the $M_i$ number of clusters for the $i$-th layer. Suppose that the weights ${\bf{W}}_{i}$ are already divided into $M_i$ clusters ${\bf{W}}_{i1}$, ${\bf{W}}_{i2}$, ..., ${\bf{W}}_{iM_i}$. Then the Euclidean projection is to set every element in ${\bf{W}}_{i}^{k+1}+{\bf{V}}_{i}^{k}$ to the average value of its cluster.
 The details of how to divide the weights into clusters will be discussed later in the next section.

After solving the subproblems, we update the dual variables ${\bf{Z}}_{i}$ and ${\bf{Y}}_{i}$, which are given by
\begin{align}
\label{U_update}
  {\bf{U}}_{i}^{k+1}&:={\bf{U}}_{i}^{k}+{\bf{W}}_{i}^{k+1}-{\bf{Z}}_{i}^{k+1},\\
\label{V_update}
  {\bf{V}}_{i}^{k+1}&:={\bf{V}}_{i}^{k}+{\bf{W}}_{i}^{k+1}-{\bf{Y}}_{i}^{k+1}.
\end{align}
This is one iteration of ADMM, and we solve the subproblems and update the dual variables iteratively until the convergence of ADMM. Namely, the following conditions need to be satisfied
\begin{align}
\label{Z_converge}
& \| {\bf{W}}_{i}^{k+1}-{\bf{Z}}_{i}^{k+1}  \|_{F}^{2} \ \le \epsilon_{i}, \ \  \| {\bf{Z}}_{i}^{k+1}-{\bf{Z}}_{i}^{k}  \|_{F}^{2} \ \le \epsilon_{i}, \\
\label{Y_converge}
& \| {\bf{W}}_{i}^{k+1}-{\bf{Y}}_{i}^{k+1}  \|_{F}^{2} \ \le \epsilon_{i}, \ \  \| {\bf{Y}}_{i}^{k+1}-{\bf{Y}}_{i}^{k}  \|_{F}^{2} \ \le \epsilon_{i}.
\end{align}

\subsection{More Understanding about the ADMM-based Framework}

An interpretation of the high performance of ADMM-based framework is as follows. It can be understood as a smart, dynamic DNN regularization technique (see Eqn. (\ref{4})), in which the regularization targets are dynamically updated in each ADMM iteration through solving of subproblems 2 and 3. This dynamic characteristics is one of the key reason that the ADMM-based framework outperforms many prior work on DNN model compression based on $L_1$ regularization (without updating of regularization targets), or Projected Gradient Descent \cite{zhang2018learning}.

\subsection{Simplification for the Proposed Framework}

The above formulation and iterative solution has high complexity. To address this issue, we present a method that prunes the unimportant weights first and then performs weight clustering or quantization. The underlying reason for this order is the higher degree of redundancy in the number of weights than the number of bits for weight representations (and therefore higher gain in weight pruning than weight clustering/quantization), as discussed before. 

In the first step, we only account for the constraints for DNN weight pruning. We update ${\bf{W}}_{i}$ and ${\bf{b}}_{i}$ according to
\begin{equation}
 \underset{ \{{\bf{W}}_{i}\},\{{\bf{b}}_{i} \}}{\text{minimize}}
\ \ \  f \big( \{{\bf{W}}_{i} \}_{i=1}^N, \{{\bf{b}}_{i} \}_{i=1}^N \big)+\sum_{i=1}^{N} \frac{\rho_{i}}{2}  \| {\bf{W}}_{i}-{\bf{Z}}_{i}^{k}+{\bf{U}}_{i}^{k} \|_{F}^{2} \\,
\end{equation}
and update ${\bf{Z}}_{i}$ and ${\bf{U}}_{i}$ according to (\ref{6}) and (\ref{U_update}).

After weight pruning, we solve DNN weight clustering or quantization problem. We consider the constraints for weight clustering or quantization on the pruned model (the remaining weights). 
To solve this problem, we update ${\bf{W}}_{i}$ and ${\bf{b}}_{i}$ according to
\begin{equation}
 \underset{ \{{\bf{W}}_{i}\},\{{\bf{b}}_{i} \}}{\text{minimize}}
\ \ \  f \big( \{{\bf{W}}_{i} \}_{i=1}^N, \{{\bf{b}}_{i} \}_{i=1}^N \big)+\sum_{i=1}^{N} \frac{\rho_{i}}{2}  \| {\bf{W}}_{i}-{\bf{Y}}_{i}^{k}+{\bf{V}}_{i}^{k} \|_{F}^{2} \\,
\end{equation}
and update ${\bf{Y}}_{i}$ and ${\bf{V}}_{i}$ according to (\ref{new_projection}) and (\ref{V_update}). Note that in weight clustering/quantization, we only update the non-zero elements. The pruned weights are fixed to zero.

The overall algorithm is shown in Algorithm 1, in which details in weight quantization/clustering will be discussed in the next section. We start from the trained DNNs (e.g., LeNet-5, AlexNet, VGGNet). We use the weight pruning ratios $\alpha_i$'s and clustering/quantization levels $M_i$'s from prior work \cite{han2015,DeepCompression} as starting points, and further increase the pruning ratios and decrease the number of clustering/quantization levels. The rationale behind this procedure is that our framework is unified and systematic and can achieve higher DNN model compression compared with state-of-the-arts.

\begin{algorithm}[h]  
\caption{ ADMM-based joint Weight Pruning and Clustering/Quantization (without details about the latter)}  
\begin{algorithmic}[1]  
    \For{{$i$ in number of layers} } 
        \State Initialize $\alpha_i$ value in every layer;  
    \EndFor
    \For{each $k$ in ADMM iterations}
      \State Solve subproblem (14) and update ${\bf{W}}_{i}$'s and ${\bf{b}}_{i}$'s;
      \For{{$i$ in the number of layers} } 
      \State Update ${\bf{Z}}_{i}$'s by performing Euclid mapping (7);
      \State Update ${\bf{U}}_{i}$'s using Eqn. (10);
      \EndFor
    \EndFor
    \State Retrain the weights that are not converged yet, and conclude the weight pruning process;
    \State Perform ADMM-based weight clustering/quantization (details to be discussed next).
\end{algorithmic}  
\end{algorithm} 

\section{Details in Weight Quantization and Clustering}
Weight clustering and quantization are used to further compress the weight representation after ADMM-based pruning. After weight pruning, there exists lots of zeros in weight matrix. We use $n$ bits, which means there are $2^n$ points representing different weights, to cluster or quantize the rest of non-zero weights (zero weights are already pruned). The number of bits and their representations can be different for different layers of DNN. The difference between weight clustering and quantization is that in weight clustering centroids of clusters are flexible, while the quantization levels (centroids) in weight quantization are fixed and predefined.

In the following we first discuss details of weight quantization, which can be perceived as a special case of weight clustering, and then the general weight clustering process.

\subsection{Details in Weight Quantization}

 \subsubsection{Parameter Initialization} In this work we use equal-distance quantization to facilitate hardware implementations. In each layer $i$ the number of quantization levels is $M_i$. We quantize the weights into a set of quantization levels $\displaystyle\{\frac{M_i}{2}q_i,  ..., 2\cdot q_i, q_i , -q_i, -2\cdot q_i, ..., -\frac{M_i}{2}q_i \}$.
The \emph{interval} $q_i$ is the distance between nearby quantization levels, which may be different for different layers. There is no need to quantize zero weights because they are already pruned.

The interval $q_i$ and quantization level $M_i$ ($n$) in weight quantization can be determined in an effective manner. For finding a value $q_i$, we denote $w_i^j$ as $j$-th weight in layer $i$ and $f(w_i^j)$ as a quantization function to the closest quantization level. Then the total square error in a single quantization step is given by $\sum_j\big|w_i^j-f(w_i^j)\big|^2$. In order to minimize the total square error, we use binary search method to determine $q_i$. To decide a value $M_i$ ($n$), we reference some prior work like \cite{han2015} and decrease $n$ accordingly. In \cite{han2015}, around 5-bit is used for quantization, which actually is a kind of clustering, in AlexNet, whereas our experiment results prove that 3-4 bits on average to quantize weights in AlexNet are sufficient without incurring any accuracy loss.

 \subsubsection{Iterative Weight Quantization and Retraining}
 After the ADMM procedure, many weights are close to the quantization levels rather than exactly on those levels, which means that we have not strictly quantized the weights yet. The reasons are twofold: the non-convexity nature of the optimization problem and the time limitation to finish the ADMM procedure. A straightforward way is to project all the weights to the nearby quantization levels. Due to the huge number of weights, although the change in every weight value is very small since they are close enough to the quantization levels, in accumulation it causes around 1\% overall accuracy degradation in our experiments (quantization to 3 bits).

To address this degradation, we present an iterative weight quantization method. In our method, we iteratively project a portion of weights to the nearby quantization levels, fix these values (i.e., we quantize these weights), and retrain the rest of them. More specifically, we quantize $\alpha$\% of weights closest to every quantization level after the ADMM procedure and then retrain the rest of weights. After we quantize the weights, we observe accuracy degradation of the DNN, while the retraining step can retrieve the accuracy. After the retraining step, we again quantize $\alpha$\% of weights closest to every quantization level and implement another retraining step. This quantization and retraining process is performed iteratively until the number of unquantized weights is small enough. Finally, we quantize these small number of remaining weights and it will not incorporate accuracy loss. 

The advantage of our proposed method is that we only quantize a portion of weights in every iteration, and our retraining step provides additional chance to the rest of the weights, so that they can be updated to retrieve the accuracy. This explains why the iterative quantization method works better than the straightforward method that quantizes all the weights directly. We show the overall algorithm about ADMM-based weight quantization and iterative retraining process by using Algorithm 2.
\begin{algorithm}[h]  
\caption{ ADMM-based Weight Quantization and Iterative Retraining Process}  
\begin{algorithmic}[1]  
    \For{{$i$ in number of layers} } 
        \State Initialize $M_i$ and $q_i$ in every layer;  
        \State Set $\displaystyle\{q_i, 2\cdot q_i, ..., \frac{M}{2}q_i, -q_i, -2\cdot q_i, ..., -\frac{M}{2}q_i \}$;
    \EndFor
    \For{each $k$ in ADMM iterations}
      \State Solve subproblem (15) and update ${\bf{W}}_{i}$'s and ${\bf{b}}_{i}$'s;
      \For{{$i$ in the number of layers} } 
      \State Update ${\bf{Y}}_{i}$'s by performing Euclid mapping (9);
      \State Update ${\bf{V}}_{i}$'s using Eqn. (11);
      \EndFor
    \EndFor
    \For{{$k$ in number of iterations} } 
        \For{{every layer} } 
        \State Quantize $\alpha$\% of weights closest to every quantization level;
        \EndFor
        \State Perform retraining on the remaining weights;
    \EndFor
    \State Quantize the rest of weights.
\end{algorithmic}  
\end{algorithm} 

\subsection{Details in Weight Clustering}

In weight clustering, we cluster the remaining weights (after weight pruning) into $M_i$ clusters, where weights in each cluster have the same value. Different from weight quantization, the centroid value $C_j$ ($1\le j\le M_i$) for each cluster $j$ is flexible, and should be optimized along with the weight clustering procedure. In the following, we discuss the details in weight clustering that are different from quantization, including weight clustering training and retraining processes. The initialization of $M_i$ ($n$) value is the same as weight quantization, and will not be discussed in details.

\subsubsection{Weight Clustering Training} The clustering centroids need to be determined together in the training procedure. For \emph{initialization}, we perform K-means clustering ($K=M_i$) and determine each centroid as the average value of associated weights. In each ADMM iteration in weight clustering training, we perform weight mapping (Euclid mapping) of ${\bf{W}}_{i}^{k+1}+{\bf{V}}_{i}^{k}$ to the nearest centroid values, and update the weights through solving Eqn. (15). Based on weight updating, we perform K-means clustering again and update each centroid value as the average value of associated weights. In this way we perform both weight clustering and centroid updating in an effective manner.

\subsubsection{Weight Clustering Retraining} 
After finishing ADMM training of weight clustering, we perform weight clustering retraining process to avoid accuracy degradation. This retraining process is not ADMM-based, but based on the basic stochastic gradient descent. In the retraining process, we perform stochastic gradient descent only on the centroid value for each cluster. In this way we maintain the same value (centroid value) for all the weights in this cluster. The retraining process will only result in accuracy enhancement instead of accuracy degradation. This retraining flexibility of centroid values is the key reason that weight clustering has higher accuracy than quantization. Algorithm 3 illustrates the whole process of ADMM-based weight clustering and retraining.

\begin{algorithm}[htb]  
\caption{ADMM-based Weight Clustering and Retraining Process.}  
\label{alg:Framwork}  
\begin{algorithmic}[1]
    \For{{$i$ in number of layers} } 
        \State Determine the $M_i$ number of clusters in every layer;
    \EndFor
     \For{each $k$ in ADMM iterations}
      \State Solve subproblem (15) and update ${\bf{W}}_{i}$'s and ${\bf{b}}_{i}$'s;
      \For{{$i$ in the number of layers} } 
      \State Update ${\bf{Y}}_{i}$'s by performing Euclid mapping (9);
      \State Update ${\bf{V}}_{i}$'s using Eqn. (11);
      \EndFor
      \For{{$i$ in the number of layers} } 
    \State Perform K-means clustering on the weights;
        \State Update centroid values based on clustering results;
    \EndFor
    \EndFor
    \For{each $j$ in times of epoch}  
      \State Retrain the centroid values based on clustering results without ADMM ;  
    \EndFor
\end{algorithmic}  
\end{algorithm} 

\begin{table*}[h]
\centering
\caption{Comparisons of model size compression ratio on the LeNet-5 model for MNIST dataset.}\label{Table:LeNetwQuantization}
\begin{tabular}{p{4.6cm}p{1.7cm}p{1cm}p{1.1cm}p{1.65cm}p{2.5cm}p{2.7cm}}
\hline
Model &  Accuracy degradation  & No. of weights  & CONV weight bits & FC weight bits & Total data size/ \quad Compress ratio & Total model size \quad (including index)/ Compress ratio \\ 
\hline
LeNet-5 Baseline & 0.0\% &  430.5K   & 32 & 32 & 1.7MB  & 1.7MB \\
\hline
Iterative pruning \cite{DeepCompression} & 0.1\% & 35.8K  & 8 & 5 & 24.2KB / 70.2$\times$ & 52.1KB / 33$\times$\\
\hline
\bf{Our Method (Clustering)} & 0.1\% & 2.57K  & 3 & 2 (3 for output layer) & 0.89KB / 1,910$\times$ &  2.73KB / 623$\times$\\
\hline
\bf{Our Method (Quantization)} & 0.2\% & 2.57K  & 3 & 2 (3 for output layer) & 0.89KB / 1,910$\times$ &  2.73KB / 623$\times$\\
\hline

\end{tabular}
\end{table*}

\begin{table*}[h]
\begin{center}
\caption{Comparisons of model size compression ratio on the AlexNet model for ImageNet dataset.}\label{Table:LeNet5AlexNetVGGwQuantization}
\begin{tabular}{p{4.6cm}p{1.7cm}p{1cm}p{1.4cm}p{1.1cm}p{2.5cm}p{2.7cm}}
\hline
Model &  Accuracy degradation  & No. of weights  & CONV weight bits & FC weight bits & Total data size/ \quad Compress ratio & Total model size \quad (including index)/ Compress ratio \\ 
\hline
AlexNet Baseline & 0.0\% &  60.9M   & 32 & 32 & 243.6MB  & 243.6MB \\
\hline
Iterative pruning \cite{DeepCompression} & 0.0\% & 6.7M  & 8 & 5 & 5.4MB / 45$\times$ & 9.0MB / 27$\times$\\
\hline
Binary quant. \cite{Leng2017} & 3.0\% & 60.9M  & 1 & 1 & 7.3MB / 32$\times$ & 7.3MB / 32$\times$\\
\hline
Ternary quant. \cite{Leng2017} & 1.8\% & 60.9M  & 2 & 2 & 15.2MB / 16$\times$ & 15.2MB / 16$\times$\\
\hline
\bf{Our Method (Clustering)} & 0.1\% & 2.47M  & 5 & 3 & 1.16MB / 210$\times$ &  2.7MB / 90$\times$\\
\hline
\bf{Our Method (Quantization)} & 0.2\% & 2.47M  & 5 & 3 & 1.16MB / 210$\times$ &  2.7MB / 90$\times$\\
\hline
\end{tabular}
\end{center}
\end{table*}

\section{Experimental Results and Discussions}
In this section, we apply the proposed joint weight pruning and weight clustering/quantization framework on LeNet-5 \cite{lecun1998} for MNIST dataset and AlexNet \cite{krizhevsky2012imagenet} for ImageNet dataset. We focus on the total compression ratio on the overall DNN model, which depends on the number of weights and the total number of bits for weight representations. Also, we make comparisons of our model compression results with the representative works on DNN weight pruning and clustering/quantization. The comparisons show that we achieve a significant improvement in DNN model compression. We implement our experiments of LeNet-5 on Tensorflow \cite{abadi2016} and AlexNet (and VGGNet) on Caffe \cite{jia2014caffe}. Our experiments are carried out on GeForce GTX 1080Ti and NVIDIA Tesla P100 GPUs.

We initialize ADMM by using the pretrained model of LeNet-5 and AlexNet. For LeNet-5, we set the penalty parameters as $\rho_{1} = \dots = \rho_{N} =  10^{-3}$ for LeNet-5 and $\rho_{1} = \dots = \rho_{N} = 1.5\times 10^{-3}$ for AlexNet. The penalty parameters we set for weight pruning and weight clustering/quantization on a network are the same.

Our codes and models are released at the link \url{http://bit.ly/2D3F0np}.

\subsection{LeNet-5 on MNIST Dataset}
We first present the weight pruning and quantization/clustering results on the LeNet-5 model. The overall results on model size compression are shown in Table 1, while the layer-wise results are shown in Table 3. For ADMM-based weight pruning alone, we achieve up to 167$\times$ weight reduction without accuracy loss, which is notably higher than the prior work such as \cite{han2015} (12$\times$), \cite{zhang2018learning} (24.1$\times$, but this is on a different model LeNet-300-100), and \cite{aghasi2017net} (45.7$\times$, with 0.5\% accuracy degradation). 

For ADMM-based joint weight pruning and quantization, we simultaneously achieve 88$\times$ weight reduction through pruning, and use an average of 2.4-bit for quantization, without accuracy loss. In terms of weight data storage, the compression ratio reaches 1,910$\times$ when comparing with the original LeNet-5. This is clearly impressive result, when considering that each MNIST sample has 784 pixels and even logistic regression has 7.84M weights (MNIST has 10 classes). When indices (required in weight pruning) are accounted for, the whole model size reduction becomes 623$\times$. We mainly compare with \cite{han2016eie} because there lacks much prior work on joint weight pruning and quantization/clustering. We can observe significant improvements in both weight pruning and quantization compared with the prior work, demonstrating the effectiveness of ADMM framework.

When weight clustering is applied, we use the same bit for CONV and FC layers as quantization, and accuracy improvement is achieved. Note that further reduction in bit representation is difficult to achieve without accuracy degradation (because FC layer is already quantized/clustered to 2-bit). Hence quantization will be sufficient if 0.1\% accuracy is not the design consideration.

\begin{table}
\centering
\caption{Layer-wise weight pruning and weight quantization/clustering results on LeNet-5}
\begin{tabular}{p{0.8cm}p{1cm}p{1.55cm}p{1.55cm}p{0.9cm}}
\hline
Layer & No. of Weights & Number of weights after prune & Percentage of weights after prune & Weight bits  \\ \hline
conv1 & 0.5K & 0.1K & 20\%  & 5\\
conv2 & 25K & 1.33K & 5.3\%  & 3 \\
fc1 & 400K & 0.8K & 0.2\%  & 2\\
fc2 & 5K & 0.35K & 7\% & 3 \\ \hline
Total & 430.5K & 2.58K & 0.6\% & 2.4\\ \hline
\end{tabular}
\end{table}

\subsection{AlexNet on ImageNet Dataset}
In this section we present the weight pruning and quantization/clustering results on the AlexNet model. The overall results on model size compression are shown in Table 2, while the layerwise results are shown in Table 4. For ADMM-based weight pruning alone, we can achieve 24.7$\times$ weight reduction without accuracy loss, which is notably higher than the prior work such as \cite{han2015} (9$\times$), \cite{yu2017compressing} (10$\times$) and \cite{dai2017} (15.7$\times$, but starting from a smaller DNN than original AlexNet). We also performed testing on VGGNet and achieve similar results on weight pruning and quantization/clustering. For example, the weight pruning ratio is 23.4$\times$ without accuracy loss. More results are abbreviated due to space limitation.

For ADMM-based joint weight pruning and quantization, we simultaneously achieve 24.7$\times$ weight reduction through pruning, and use an average of 3.4-bit (not accounting for the first and last layers, similar to prior work) for weight quantization, without accuracy loss. In terms of weight data storage, the compression ratio reaches 210$\times$ when comparing with the original AlexNet model, which is also significant improvement. When indices in weight pruning are accounted for, the whole model size reduction becomes 90$\times$. When comparing with \cite{han2015}, we observe significant improvements in both weight pruning and quantization compared with the prior work, demonstrating the effectiveness of the ADMM-based framework.

Finally when it comes to weight clustering, we use the same bit for CONV and FC layers as quantization, and accuracy improvement is observed. Again due to the difficulty for further bit representation reduction without accuracy loss, weight quantization will be sufficient for most of the application domains.

\begin{table}
\centering
\caption{Layer-wise weight pruning and weight quantization/clustering results on AlexNet}
\begin{tabular}{p{0.8cm}p{1cm}p{1.55cm}p{1.55cm}p{0.9cm}}
\hline
Layer & No. of Weights & Number of weights after prune & Percentage of weights after prune & Weight bits  \\ \hline
conv1 & 34.8K & 28.19K & 81\% &  8\\
conv2 & 307.2K & 61.44K & 20\% &  5\\
conv3 & 884.7K & 168.09K & 19\% &  5\\
conv4 & 663.5K & 132.7K & 20\% &  5\\
conv5 & 442.4K & 88.48K & 20\% &  5\\
fc1 & 37.7M & 0.75M & 2\% & 3 \\
fc2 & 16.8M & 0.91M & 5.4\% &  3\\
fc3 & 4.1M & 0.33M & 8\% &  8\\ \hline
Total & 60.9M & 2.47M & 4.06\% & 3.4\\ \hline
\end{tabular}
\end{table}

\section{Conclusion}
In this paper, we present a unified framework of DNN weight pruning and weight clustering/quantization using ADMM. When we focus on weight pruning alone, we achieve 167$\times$ weight reduction in LeNet-5, 24.7$\times$ in AlexNet, and 23.4$\times$ in VGGNet without accuracy loss. For the combination of DNN weight pruning and clustering/quantization, we achieve 1,910$\times$ and 210$\times$ storage reduction of weight data on LeNet-5 and AlexNet, respectively, without accuracy loss.


\begin{thebibliography}{}

\bibitem[\protect\citeauthoryear{Abadi \bgroup et al\mbox.\egroup
  }{2016}]{abadi2016}
Abadi, M.; Agarwal, A.; Barham, P.; et~al.
\newblock 2016.
\newblock Tensorflow: Large-scale machine learning on heterogeneous distributed
  systems.
\newblock {\em arXiv preprint arXiv:1603.04467}.

\bibitem[\protect\citeauthoryear{Aghasi \bgroup et al\mbox.\egroup
  }{2017}]{aghasi2017net}
Aghasi, A.; Abdi, A.; Nguyen, N.; and Romberg, J.
\newblock 2017.
\newblock Net-trim: Convex pruning of deep neural networks with performance
  guarantee.
\newblock In {\em Advances in Neural Information Processing Systems},
  3177--3186.

\bibitem[\protect\citeauthoryear{Boyd \bgroup et al\mbox.\egroup
  }{2011}]{boyd2011}
Boyd, S.; Parikh, N.; Chu, E.; Peleato, B.; and Eckstein, J.
\newblock 2011.
\newblock Distributed optimization and statistical learning via the alternating
  direction method of multipliers.
\newblock {\em Foundations and Trends{\textregistered} in Machine Learning}
  3(1):1--122.

\bibitem[\protect\citeauthoryear{Cheng \bgroup et al\mbox.\egroup
  }{2015}]{cheng2015exploration}
Cheng, Y.; Yu, F.~X.; Feris, R.~S.; Kumar, S.; Choudhary, A.; and Chang, S.-F.
\newblock 2015.
\newblock An exploration of parameter redundancy in deep networks with
  circulant projections.
\newblock In {\em Proceedings of the IEEE International Conference on Computer
  Vision},  2857--2865.

\bibitem[\protect\citeauthoryear{Dai, Yin, and Jha}{2017}]{dai2017}
Dai, X.; Yin, H.; and Jha, N.~K.
\newblock 2017.
\newblock Nest: A neural network synthesis tool based on a grow-and-prune
  paradigm.
\newblock {\em arXiv preprint arXiv:1711.02017}.

\bibitem[\protect\citeauthoryear{Guo, Yao, and Chen}{2016}]{guo2016dynamic}
Guo, Y.; Yao, A.; and Chen, Y.
\newblock 2016.
\newblock Dynamic network surgery for efficient dnns.
\newblock In {\em Advances In Neural Information Processing Systems},
  1379--1387.

\bibitem[\protect\citeauthoryear{Han \bgroup et al\mbox.\egroup
  }{2015}]{han2015}
Han, S.; Pool, J.; Tran, J.; and Dally, W.
\newblock 2015.
\newblock Learning both weights and connections for efficient neural network.
\newblock In {\em Advances in Neural Information Processing Systems (NIPS)},
  1135--1143.

\bibitem[\protect\citeauthoryear{Han \bgroup et al\mbox.\egroup
  }{2016}]{han2016eie}
Han, S.; Liu, X.; Mao, H.; Pu, J.; Pedram, A.; Horowitz, M.~A.; and Dally,
  W.~J.
\newblock 2016.
\newblock Eie: efficient inference engine on compressed deep neural network.
\newblock In {\em Computer Architecture (ISCA), 2016 ACM/IEEE 43rd Annual
  International Symposium on},  243--254.
\newblock IEEE.

\bibitem[\protect\citeauthoryear{Han, Mao, and Dally}{2016}]{DeepCompression}
Han, S.; Mao, H.; and Dally, W.~J.
\newblock 2016.
\newblock Deep compression: Compressing deep neural networks with pruning,
  trained quantization and huffman coding.
\newblock {\em International Conference on Learning Representations (ICLR)}.

\bibitem[\protect\citeauthoryear{Jia \bgroup et al\mbox.\egroup
  }{2014}]{jia2014caffe}
Jia, Y.; Shelhamer, E.; Donahue, J.; Karayev, S.; Long, J.; Girshick, R.;
  Guadarrama, S.; and Darrell, T.
\newblock 2014.
\newblock Caffe: Convolutional architecture for fast feature embedding.
\newblock In {\em Proceedings of the 22nd ACM international conference on
  Multimedia},  675--678.
\newblock ACM.

\bibitem[\protect\citeauthoryear{Kingma and Ba}{2014}]{kingma2014adam}
Kingma, D.~P., and Ba, J.
\newblock 2014.
\newblock Adam: A method for stochastic optimization.
\newblock {\em arXiv preprint arXiv:1412.6980}.

\bibitem[\protect\citeauthoryear{Krizhevsky, Sutskever, and
  Hinton}{2012}]{krizhevsky2012imagenet}
Krizhevsky, A.; Sutskever, I.; and Hinton, G.~E.
\newblock 2012.
\newblock Imagenet classification with deep convolutional neural networks.
\newblock In {\em Advances in neural information processing systems},
  1097--1105.

\bibitem[\protect\citeauthoryear{LeCun \bgroup et al\mbox.\egroup
  }{1998}]{lecun1998}
LeCun, Y.; Bottou, L.; Bengio, Y.; and Haffner, P.
\newblock 1998.
\newblock Gradient-based learning applied to document recognition.
\newblock {\em Proceedings of the IEEE} 86(11):2278--2324.

\bibitem[\protect\citeauthoryear{Leng \bgroup et al\mbox.\egroup
  }{2017}]{Leng2017}
Leng, C.; Li, H.; Zhu, S.; and Jin, R.
\newblock 2017.
\newblock Extremely low bit neural network: Squeeze the last bit out with admm.
\newblock {\em arXiv preprint arXiv:1707.09870}.

\bibitem[\protect\citeauthoryear{Park, Ahn, and Yoo}{2017}]{park2017weighted}
Park, E.; Ahn, J.; and Yoo, S.
\newblock 2017.
\newblock Weighted-entropy-based quantization for deep neural networks.
\newblock In {\em IEEE Conference on Computer Vision and Pattern Recognition
  (CVPR)}.

\bibitem[\protect\citeauthoryear{Sindhwani, Sainath, and
  Kumar}{2015}]{sindhwani2015structured}
Sindhwani, V.; Sainath, T.; and Kumar, S.
\newblock 2015.
\newblock Structured transforms for small-footprint deep learning.
\newblock In {\em Advances in Neural Information Processing Systems},
  3088--3096.

\bibitem[\protect\citeauthoryear{Takapoui \bgroup et al\mbox.\egroup
  }{2017}]{takapoui2017simple}
Takapoui, R.; Moehle, N.; Boyd, S.; and Bemporad, A.
\newblock 2017.
\newblock A simple effective heuristic for embedded mixed-integer quadratic
  programming.
\newblock {\em International Journal of Control}  1--11.

\bibitem[\protect\citeauthoryear{Wen \bgroup et al\mbox.\egroup
  }{2016}]{wen2016learning}
Wen, W.; Wu, C.; Wang, Y.; Chen, Y.; and Li, H.
\newblock 2016.
\newblock Learning structured sparsity in deep neural networks.
\newblock In {\em Advances in Neural Information Processing Systems},
  2074--2082.

\bibitem[\protect\citeauthoryear{Wen \bgroup et al\mbox.\egroup
  }{2017}]{wen2017coordinating}
Wen, W.; Xu, C.; Wu, C.; Wang, Y.; Chen, Y.; and Li, H.
\newblock 2017.
\newblock Coordinating filters for faster deep neural networks.
\newblock {\em CoRR, abs/1703.09746}.

\bibitem[\protect\citeauthoryear{Yang, Chen, and Sze}{2016}]{yang2016}
Yang, T.-J.; Chen, Y.-H.; and Sze, V.
\newblock 2016.
\newblock Designing energy-efficient convolutional neural networks using
  energy-aware pruning.
\newblock {\em arXiv preprint arXiv:1611.05128}.

\bibitem[\protect\citeauthoryear{Ye \bgroup et al\mbox.\egroup
  }{2018}]{ye2018progressive}
Ye, S.; Zhang, T.; Zhang, K.; Li, J.; Xu, K.; Yang, Y.; Yu, F.; Tang, J.;
  Fardad, M.; Liu, S.; et~al.
\newblock 2018.
\newblock Progressive weight pruning of deep neural networks using admm.
\newblock {\em arXiv preprint arXiv:1810.07378}.

\bibitem[\protect\citeauthoryear{Yu \bgroup et al\mbox.\egroup
  }{2017}]{yu2017compressing}
Yu, X.; Liu, T.; Wang, X.; and Tao, D.
\newblock 2017.
\newblock On compressing deep models by low rank and sparse decomposition.
\newblock In {\em Proceedings of the IEEE Conference on Computer Vision and
  Pattern Recognition},  7370--7379.

\bibitem[\protect\citeauthoryear{Zhang \bgroup et al\mbox.\egroup
  }{2018a}]{zhang2018learning}
Zhang, D.; Wang, H.; Figueiredo, M.; and Balzano, L.
\newblock 2018a.
\newblock Learning to share: Simultaneous parameter tying and sparsification in
  deep learning.

\bibitem[\protect\citeauthoryear{Zhang \bgroup et al\mbox.\egroup
  }{2018b}]{zhang2018systematic}
Zhang, T.; Ye, S.; Zhang, K.; Tang, J.; Wen, W.; Fardad, M.; and Wang, Y.
\newblock 2018b.
\newblock A systematic dnn weight pruning framework using alternating direction
  method of multipliers.
\newblock {\em arXiv preprint arXiv:1804.03294}.

\bibitem[\protect\citeauthoryear{Zhang \bgroup et al\mbox.\egroup
  }{2018c}]{zhang2018adam}
Zhang, T.; Zhang, K.; Ye, S.; Li, J.; Tang, J.; Wen, W.; Lin, X.; Fardad, M.;
  and Wang, Y.
\newblock 2018c.
\newblock Adam-admm: A unified, systematic framework of structured weight
  pruning for dnns.
\newblock {\em arXiv preprint arXiv:1807.11091}.

\bibitem[\protect\citeauthoryear{Zhao \bgroup et al\mbox.\egroup
  }{2017}]{zhao2017theoretical}
Zhao, L.; Liao, S.; Wang, Y.; Li, Z.; Tang, J.; Pan, V.; and Yuan, B.
\newblock 2017.
\newblock Theoretical properties for neural networks with weight matrices of
  low displacement rank.
\newblock {\em arXiv preprint arXiv:1703.00144}.

\bibitem[\protect\citeauthoryear{Zhou \bgroup et al\mbox.\egroup
  }{2016}]{zhou2016dorefa}
Zhou, S.; Wu, Y.; Ni, Z.; Zhou, X.; Wen, H.; and Zou, Y.
\newblock 2016.
\newblock Dorefa-net: Training low bitwidth convolutional neural networks with
  low bitwidth gradients.
\newblock {\em arXiv preprint arXiv:1606.06160}.

\bibitem[\protect\citeauthoryear{Zhou \bgroup et al\mbox.\egroup
  }{2017}]{zhou2017incremental}
Zhou, A.; Yao, A.; Guo, Y.; Xu, L.; and Chen, Y.
\newblock 2017.
\newblock Incremental network quantization: Towards lossless cnns with
  low-precision weights.
\newblock {\em arXiv preprint arXiv:1702.03044}.

\end{thebibliography}

\end{document}